\documentclass{article}

\usepackage{microtype}
\usepackage{graphicx}
\usepackage{subfigure}
\usepackage{booktabs} % for professional tables

\usepackage{hyperref}
\usepackage{algpseudocode}

\usepackage[accepted]{icml2025}

\usepackage{amsmath}
\usepackage{amssymb}
\usepackage{mathtools}
\usepackage{amsthm}

\usepackage[capitalize,noabbrev]{cleveref}

\theoremstyle{plain}

\theoremstyle{definition}

\theoremstyle{remark}

\usepackage[textsize=tiny]{todonotes}

\usepackage{adjustbox}
\usepackage{fixltx2e}
\usepackage{makecell}
\usepackage{wrapfig}
\usepackage{amssymb}
\usepackage{enumitem}
\usepackage{graphicx}
\usepackage{caption}     % better‑looking captions and spacing

\newcommand{\E}{\mathbb{E}}

\newcommand{\modelname}[1]{$\small{\text{TABASCO}}$}
\newcommand{\bigmodelname}[1]{$\large{\text{TABASCO}}$}
\newcommand{\posebusters}[1]{\textsc{PoseBusters}}
\icmltitlerunning{\modelname{}: A Fast, Simplified Model for Molecular Generation with Improved Physical Quality}

\begin{document}

\twocolumn[
\icmltitle{TABASCO: A Fast, Simplified Model for Molecular Generation\\with Improved Physical Quality}

% It is OKAY to include author information, even for blind
% submissions: the style file will automatically remove it for you
% unless you've provided the [accepted] option to the icml2025
% package.

% List of affiliations: The first argument should be a (short)
% identifier you will use later to specify author affiliations
% Academic affiliations should list Department, University, City, Region, Country
% Industry affiliations should list Company, City, Region, Country

% You can specify symbols, otherwise they are numbered in order.
% Ideally, you should not use this facility. Affiliations will be numbered
% in order of appearance and this is the preferred way.
\icmlsetsymbol{core}{*}
\icmlsetsymbol{visitor}{\dag}

\begin{icmlauthorlist}
\icmlauthor{Carlos Vonessen}{core,yyy,visitor}
\icmlauthor{Charles Harris}{core,zzz}
\icmlauthor{Miruna Cretu}{core,zzz}
\icmlauthor{Pietro Li\`{o}}{zzz}
\end{icmlauthorlist}

\icmlaffiliation{yyy}{ETH Zurich, Switzerland}
\icmlaffiliation{zzz}{University of Cambridge, United Kingdom}
% \icmlaffiliation{aaa}{Work done while visiting at the University of Cambridge}

\icmlcorrespondingauthor{Carlos Vonessen}{cvonessen[at]ethz.ch}
\icmlcorrespondingauthor{Charles Harris}{cch57[at]cam.ac.uk}

% You may provide any keywords that you
% find helpful for describing your paper; these are used to populate
% the "keywords" metadata in the PDF but will not be shown in the document
\icmlkeywords{Molecule Generation, Diffusion, Machine Learning}

\vskip 0.3in
]

% this must go after the closing bracket ] following \twocolumn[ ...

% This command actually creates the footnote in the first column
% listing the affiliations and the copyright notice.
% The command takes one argument, which is text to display at the start of the footnote.
% The \icmlEqualContribution command is standard text for equal contribution.
% Remove it (just {}) if you do not need this facility.

\printAffiliationsAndNotice{\textsuperscript{*}Core contributor, \textsuperscript{\dag} Work done as a visiting student in Cambridge}  % leave blank if no need to mention equal contribution
% \printAffiliationsAndNotice{\icmlEqualContribution} % otherwise use the standard text.

\begin{abstract}
State-of-the-art models for 3D molecular generation are based on significant inductive biases—\textit{SE}(3), permutation equivariance to respect symmetry and graph message‑passing networks to capture local chemistry—yet the generated molecules still struggle with physical plausibility.
We introduce \modelname{} which relaxes these assumptions: The model has a standard non-equivariant transformer architecture, treats atoms in a molecule as sequences and reconstructs bonds deterministically after generation. The absence of equivariant layers and message passing allows us to significantly simplify the model architecture and scale data throughput.
On the GEOM‑Drugs benchmark \modelname{} achieves state-of-the-art PoseBusters validity and delivers inference roughly $10\times$ faster than the strongest baseline, while exhibiting emergent rotational equivariance despite symmetry not being hard‑coded. 
Our work offers a blueprint for training minimalist, high‑throughput generative models suited to specialised tasks such as structure‑ and pharmacophore‑based drug design.
We provide a link to our implementation at \href{github.com/carlosinator/tabasco}{\texttt{github.com/carlosinator/tabasco}}.
\end{abstract}

\begin{figure}
    % \centering
    \includegraphics[width=0.9\linewidth]{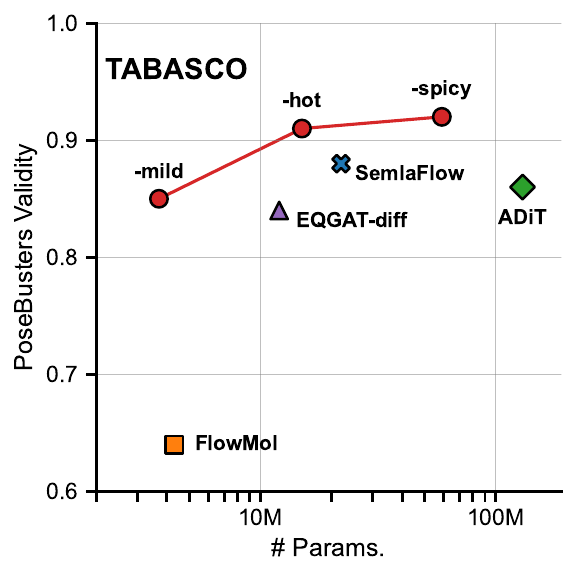}
    \caption{Comparison of \posebusters{} on GEOM-Drugs. \vspace{-15pt}}
    \label{fig:pareto}
\end{figure}

\section{Introduction}

In recent years, there has been growing interest in using diffusion models as generative methods for molecular design \cite{du2024review1,schneuing2022diffsbdd, pmlr-v162-hoogeboom22a, vignac2023midi,dunn2024flowmol, irwin2025semlaflow}. Much of the literature converges on design principles believed to be essential for high-quality molecular generation. First, models are typically SE(3)-equivariant, ensuring that rotations or translations of input conformers yield correspondingly transformed outputs—a symmetry prior that serves as a strong inductive bias \cite{pmlr-v162-hoogeboom22a}. Second, message-passing graph neural networks (GNNs) are widely used to capture many-hop, context-dependent interactions between atoms \cite{pmlr-v162-hoogeboom22a, schneuing2022diffsbdd, irwin2025semlaflow, dunn2024flowmol, schneuingdrugflow}. Third, recent work emphasises flow-matching objectives that rely on coupled optimal transport (OT) \cite{tong2023improving} or incorporate heavily structured, domain-informed priors \cite{dunn2024flowmol, irwin2025semlaflow}. However, despite incorporating these inductive biases, current models continue to struggle with physical plausibility—often failing to produce chemically coherent structures or accurately recover fundamental features of protein–ligand binding \citep{buttenschoen2024posebusters, harris2023posecheck}.

In parallel, a growing body of work explores scaling up simpler model architectures—most notably Transformers \cite{vaswani2017attention}—across adjacent domains. A prominent example is AlphaFold3 \citep{abramson2024alphafold3}, which achieves strong performance on physical plausibility benchmarks \citep{buttenschoen2024posebusters} despite omitting many of the conventional inductive biases, including equivariance. Similarly, recent generative models for protein backbone design have demonstrated competitive results with minimal architectural complexity, provided they are scaled appropriately \citep{geffner2025proteina}. Simplification of model architecture and removal of inductive biases for conformer generation has also proven successful \cite{wang2024swallowingbitterpill}. In parallel to this work, \cite{joshi2025allatomdiffusiontransformer} explore using non-equivariant latent diffusion for generating small molecules.

In this work, we aim to distill the core components of diffusion-based molecular generation and ask: how much architectural complexity is necessary to build high-performing models? We introduce \modelname{} (Transformer-based Atomistic Bondless Scalable Conformer Output), a stripped-down and scalable model that achieves state-of-the-art performance on unconditional molecular generation benchmarks. Despite its simplicity, \modelname{} exceeds the physical plausibility of more complex models, as measured by \posebusters{} \cite{buttenschoen2024posebusters} validity, while being up to 10$\times$ faster at inference. Our contributions are as follows:

\begin{enumerate}[leftmargin=*,label=(\roman*)]

    \item \textbf{State-of-the-art physical quality on GEOM-Drugs.}
    \modelname{} surpasses prior models such as FlowMol and SemlaFlow in \posebusters{} validity, achieving a 10$\times$ speed-up at sampling time.
    
    \item \textbf{Lean, bond-free Transformer backbone.}  
    Our model omits both bond inputs and equivariant layers, relying instead on a standard Transformer to generate high-quality coordinates. Chemoinformatics tools recover bonds post hoc, which maintains physical plausibility and focuses computational resources on exact coordinate generation.

    \item \textbf{Physically-constrained last-mile correction.}  
    We introduce a simple distance-bounds guidance step that improves \textsc{PoseBusters} validity without requiring force-field-based relaxation or additional parameters.
    
    \item \textbf{Emergent structure without explicit symmetry.}  
    We analyse the model’s equivariant behaviour despite the absence of SE(3) symmetry constraints, and investigate the role of positional encodings in improving model performance.
\end{enumerate}

\section{Background and Related Work}

\subsection{Flow-Matching Models}
% general intro to Diffusion and CFM

Flow-matching (FM) is a generative modelling framework that learns to transport samples from a source distribution (e.g., noise) to a target distribution (e.g., data) by directly estimating the time-dependent velocity field of a probability flow \cite{lipman2023flowmatchinggenerativemodeling, albergo2022building}.

Given a pair of samples $(\mathbf{x}_0, \mathbf{x}_1)$ from source and target distributions, one defines a continuous interpolation $\mathbf{x}_t = (1 - t)\mathbf{x}_0 + t\mathbf{x}_1$, and a target velocity $u_t = \frac{\mathbf{x}_1 - \mathbf{x}0}{t(1 - t)}$. A neural field $v_\theta(\mathbf{x}_t, t)$ is then trained to match this velocity using the squared error:
\begin{equation}
\mathcal{L}_{\text{FM}} = \mathbb{E}_{t, (\mathbf{x}_0, \mathbf{x}1)} \left[ || v_\theta(\mathbf{x}_t, t) - u_t ||_2^2 \right].
\end{equation}

Flow-matching enables efficient generation via deterministic integration (e.g., using an ODE solver), and has been shown to improve sampling speed and stability over score-based diffusion models \cite{dunn2024flowmol, irwin2025semlaflow}.

\subsection{Generative Models for 3D Molecule Design}

Early works used standard continuous diffusion processes on coordinate and atomic features, where bond connectivity was determined by chemoinformatics software \cite{pmlr-v162-hoogeboom22a, schneuing2022diffsbdd}. This process often resulted in low-quality conformers that were not fully-connected or violated atomic valences.
% Midi
MiDi \cite{vignac2023midi} improved on this by applying discrete diffusion to both the atom types as well as generated a full bond matrix end-to-end, which significantly increased stabilty and bond connectivity.
% EQGAT-diff
EQGAT-diff \cite{le2023eqgat} explored the design space of equivariant diffusion models, creating a custom attention-based equivariant architecture to allow for interaction between continuous and discrete features.
% further work
Further work introduced more advanced model architectures \cite{morehead2024geometrycompletediffusion3dmolecule, hua2024mudiffunifieddiffusioncomplete}, additional losses \cite{Xu_2024}, alternative transport strategies \cite{song2023equivariantflowmatchinghybrid}, and geometric latent diffusion \cite{xu2023geometriclatentdiffusionmodels, joshi2025allatomdiffusiontransformer}.
% FlowMol and Semla Flow
FlowMol \cite{dunn2024flowmol} and SemlaFlow \cite{irwin2025semlaflow} use flow-matching for generation of coordinates, atom types and bonds. Both methods proposed new architectures and showed great improvements in speed versus diffusion based approaches.

\section{\bigmodelname{}: Fast, Simple, and High-Quality Molecule Generation}

\begin{figure*}
    \centering
    \includegraphics[width=\textwidth]{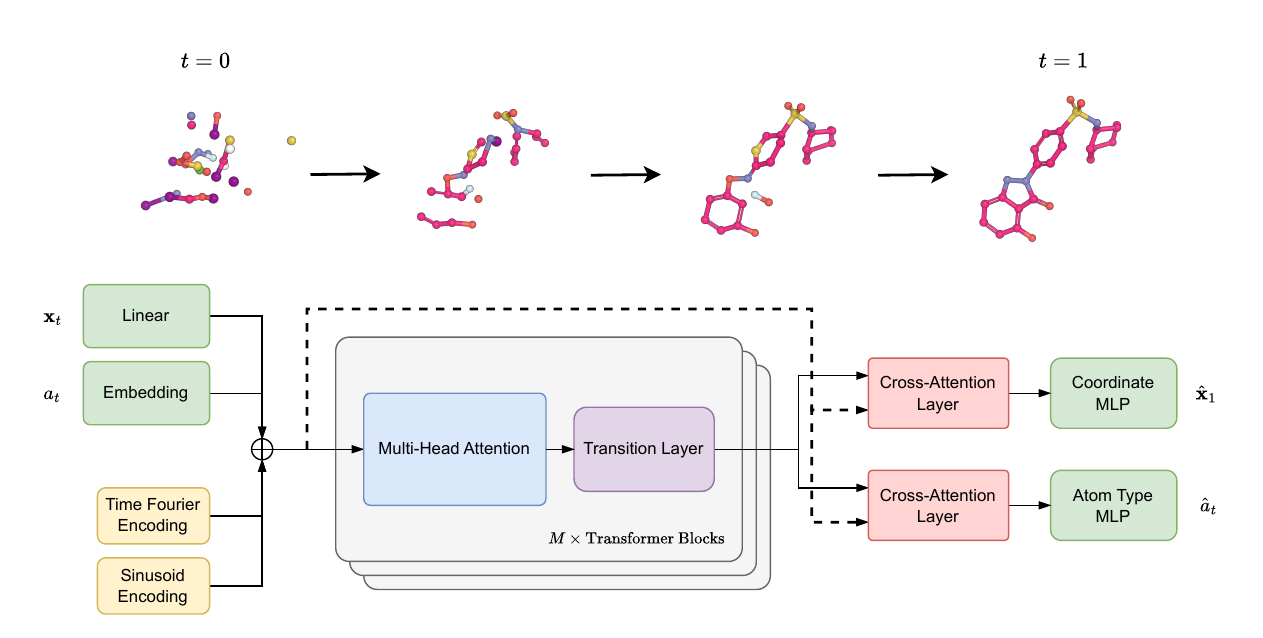}
    \caption{Top: Interpolation between noise and data. Bottom: \modelname{} model architecture.}
    \label{fig:model-arch}
\end{figure*}

\paragraph{Overview and Motivation}

Our goal in this work is to identify the simplest possible model architecture that can generate physically realistic small molecules at scale. Our motivation stems from the observation that recent progress in protein structure generation has demonstrated the surprising power of non-equivariant Transformer architectures when scaled appropriately \cite{abramson2024alphafold3, geffner2025proteina, wang2024swallowingbitterpill}. Based on these results, we began our experiments with a deliberately stripped-down, non-equivariant Transformer backbone for molecular generation.

We choose to exclude explicit bond information from the model. While most existing models treat bonds as a distinct modality, often processed with triangle attention or edge representations, we rely on standard chemoinformatics tools instead, which infer bonds reliably so long as the generated coordinates are physically sensible. We therefore hypothesised that if coordinate generation is sufficiently accurate, bond information becomes redundant. This perspective allowed us to further simplify the architecture while focusing on improving conformer quality.\\
Physical realism, as measured by \textsc{PoseBusters} validity, is the primary metric guiding design decisions. Modules and heuristics in the approach that did not contribute to this metric were pruned, resulting in a lean, fast, and extensible model that maintains strong performance without relying on specialised architectural components.

\subsection{Model Architecture}
% omg, its so simple, easy, and extensible!

In contrast to most prior work in unconditional molecular generation, we adopt a simplified non-equivariant Transformer architecture (see Figure \ref{fig:model-arch}) without self-conditioning. Atom coordinates and types are jointly embedded along with time and sequence encodings. These are passed through a stack of standard Transformer blocks \cite{vaswani2017attention}. We add a single cross-attention layer for each domain and process these outputs in MLP heads for atom types and coordinates. The resulting model is straightforward to implement, highly extensible, and dramatically faster at sampling time than previous equivariant or bond-aware approaches.

\subsection{Training Objective}
We optimize coordinates with Euclidean conditional flow-matching (CFM) \cite{tong2023improving, albergo2022building} and atom types with discrete CFM which is parametrized based on the Discrete Flow Models (DFM) framework \cite{campbell}. Concretely, consider a molecule with $N$ atoms, ground-truth coordinates $\mathbf{x_1}$ and atom types $a_1$. Coordinates are partially noised with $\mathbf{x_t}=t\cdot \mathbf{x_1} + (1-t)\cdot\mathbf{\epsilon}$, where the noise is distributed with $\mathbf{\epsilon}\sim\mathcal{N}(0, I)$. Noisy atom types $a_t$ are obtained by interpolating between atom type probabilities and sampling from a cate\-gorical distribution $a_t\sim\text{Cat}\left(t\cdot \delta(a_1) + (1-t)\cdot\delta(\frac{1}{N})\right)$, where $\delta(\cdot)$ creates a one-hot encoding \cite{campbell}. During training the model takes $\mathbf{x_t}$ and $a_t$ and learns to predict the endpoint of the trajectory. The continuous coordinate objective becomes

\begin{equation}
    L_{\text{metric}}(\mathbf{x}) =\E_{\epsilon,t}\left[\frac{1}{N}||\hat{\mathbf{x}}_1^{\theta}(\mathbf{x}_t, t) - \mathbf{x}_1||_2^2\right]\,.
\end{equation}

The discrete atom type objective is the cross-entropy loss

\begin{equation}
    L_{\text{discrete}}(a)=\E_{t}\left[ - \sum_{i} a_i \log(\hat{a}_1(a_t, t))\right]\,.
\end{equation}

We combine these into a multi-objective formulation with weighing factor $\lambda_\text{discrete}\in(0, 1]$, as

\begin{equation}
    L_\text{total}(\mathbf{x}, a) = L_{\text{metric}}(\mathbf{x}) + \lambda_\text{discrete}\cdot L_{\text{discrete}}(a) \,.
\end{equation}

During training we sample from $t\sim\text{Beta}(\alpha,1)$, where $\alpha$ is a hyperparameter we ablate in Appendix~\ref{app:sampling}. As $t \rightarrow 1$ the model's behaviour approaches the identity function, due to the chosen endpoint formulation. To ensure the model can still learn precise atom placement even as losses approach zero as $t \rightarrow1$, we weigh the loss with $\beta(t) \cdot L_\text{total}(\mathbf{x}_t, a_t)$ based on the sampled time $t$, with

\begin{equation}
    \beta(t)=\min\left\{100, \frac{1}{(1-t)^2}\right\}\,.
\end{equation}

\begin{figure*}[t]
    \centering
    \includegraphics[width=\textwidth]{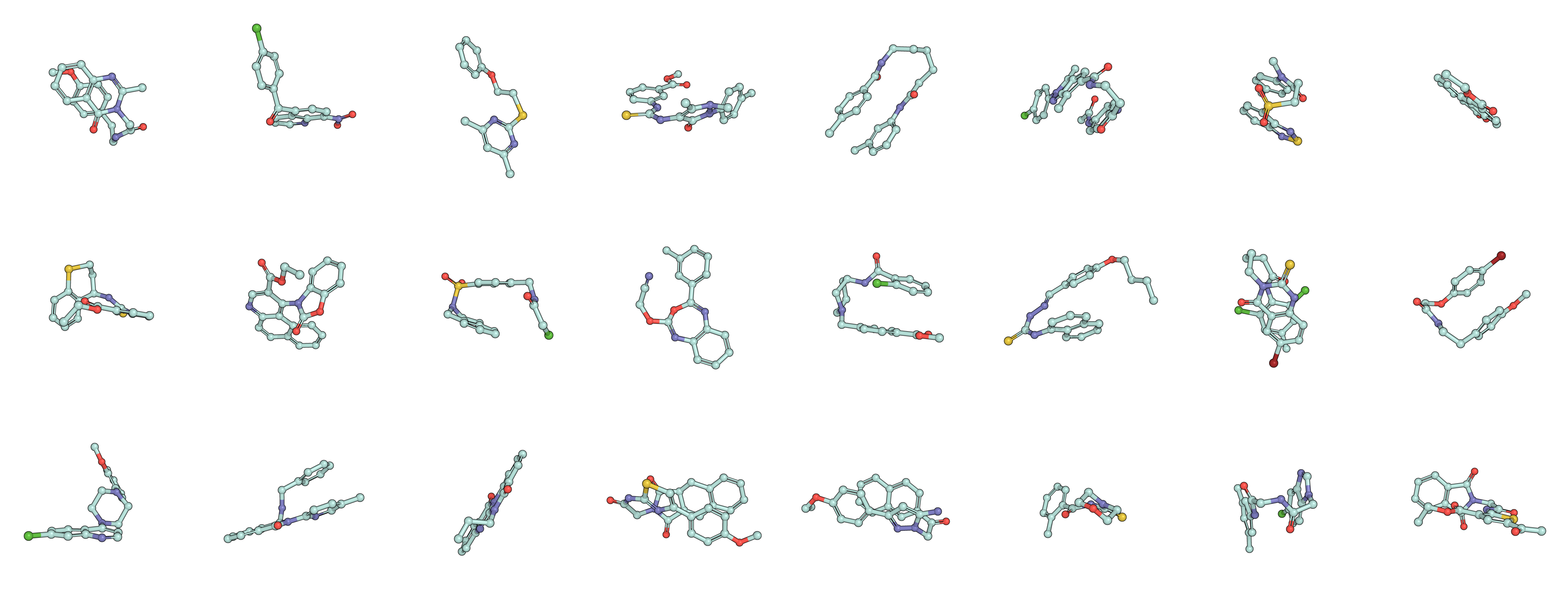}
    \caption{\textbf{Sampled molecules from \modelname{}}.}
    \label{fig:sampled-mols}
\end{figure*}

\subsection{Sampling}

We generate molecules with \modelname{} by simulating a system of coupled stochastic differential equations:
\begin{align}
\mathrm{d}\mathbf{x}_t &= \mathbf{v}_t^\theta(\mathbf{x}_t, a_t)\mathrm{d}t + g(t)\,\mathbf{s}_t^\theta(\mathbf{x}_t, a_t)\mathrm{d}t \nonumber\\
&\quad + \sqrt{2g(t)\gamma}\,\mathrm{d}W_t\;,\\
\partial p_t &= R_t(\mathbf{x}_t, a_t)^\top p_t\;
\label{eq:stochastic_sampling}
\end{align}

where $p_t$ describes the probability of each atom type at time $t$. We estimate the velocity with  $\mathbf{v}_t = \frac{\mathbf{x}_1 - \mathbf{x}_t}{1-t}$ from the models endpoint prediction $\hat{\mathbf x}_1$ at time $t$, and the score with $\mathbf{s}_t = \frac{t\mathbf{v}_t - \mathbf{x}_t}{1-t}$. We refer the reader to \cite{geffner2025proteina, campbell} on which we base our coordinate and atom type sampling strategies, for more in-detail discussions. To improve sample quality, we apply a logarithmic discretization scheme on $t\in[0, 1]$ with more fine-grained steps near the end of denoising. We also scale the score $\mathbf{s_t}$ and the Gaussian noise component $\mathrm{d}W_t$ by $g(t)$, setting it to zero as $t\rightarrow 1$ (see Appendix~\ref{app:sampling}).

\subsection{Ordering Atoms as Sequences}
\label{sec:atom_order}

Transformers operate in a bag-of-tokens fashion unless provided with additional information about the absolute or relative positions of those tokens. Unlike text or protein sequences, small molecules lack a natural linear ordering that reflects their 3D structure. While formats such as SMILES and InChI offer consistent ways to linearise molecular graphs, the ordering in these representations does not strictly correspond to spatial proximity. However, the SMILES ordering is deterministically derived—typically via a depth-first traversal starting from a canonical root atom \cite{weininger1989smiles}—which does impart some semantic structure. In practice, many neighbouring atoms in the SMILES string are also spatially or chemically proximate in the molecule. We hypothesise that this implicit locality helps the model establish a coarse structural scaffold early in the generation process (see lower trajectory in Figure~\ref{fig:two-trajectories}). Accordingly, we include sinusoidal positional encodings based on the atom indices in the SMILES sequence, and ablate their effect in Section~\ref{sec:sequential-ordering}.

\subsection{Physically Constrained Last‑Mile Pose Guidance}
\label{sec:phys_guidance}

Existing 3D molecule generators yield globally sound conformations but struggle with local stereochemical checks such as \textsc{PoseBusters}.  
We find most violations stem from subtle coordinate drifts that accumulate near the end of the sampling trajectory (\(t\!\rightarrow\!1\)).
We therefore frame pose refinement as a \emph{last‑mile} problem and introduce a lightweight, differentiable guidance step that enforces simple physical distance bounds without force‑field evaluation or relaxation.

\paragraph{Distance–bounds matrix.}
For every element pair we pre‑compute lower and upper bounds \(\bigl[L_{ij},\,U_{ij}\bigr]\) over 1–5 bond separations, analogously to how \posebusters{} computes bounds on valid bond lengths and angles:
\begin{itemize}
    \item \textbf{Lower bound \(L_{ij}\):} sum of van‑der‑Waals radii minus 0.1 Å;
    \item \textbf{Upper bound \(U_{ij}\):} cumulative covalent bond lengths along the shortest path.
\end{itemize}
These numbers match the limits used in Universal Force Field (UFF) relaxation but are looked up from a static table; \emph{no} UFF energy, gradients, or optimisation is performed. 

\paragraph{Two‑phase sampling with distance‑bounds guidance.}
\begin{enumerate}[leftmargin=1em]
    \item \textbf{Free denoising.}  Run the standard sampler until \(t = 0.99\), obtaining noised conformation \((\mathbf x_{0.99}, a_{0.99})\).
    \item \textbf{Guided refinement.}  In each remaining denoising step, convert the endpoint predicted coordinates to an \textsc{RDKit} conformer and look up the physical bounds on atom pair distances $[L_{ij},\; U_{ij}]$ for each distance pair $d_{ij}=||\mathbf{x}_{t,i}-\mathbf{x}_{t,j}||$. The loss on physical constraints is computed with
    \[
        \mathcal{L}_{\text{phys}}(\mathbf{x}_t)=
        \sum_{i<j}
        \begin{cases}
            \bigl(d_{ij}-U_{ij}\bigr)^{2}, & d_{ij}>U_{ij},\\[4pt]
            \bigl(L_{ij}-d_{ij}\bigr)^{2}, & d_{ij}<L_{ij},\\[4pt]
            0, & \text{otherwise}\,.
        \end{cases}
    \]
    We back‑propagate through the network and apply one gradient step to the inputs:
    \[
        \mathbf{x}_t  \;\leftarrow\; \mathbf{x}_t-\alpha_{\text{phys}}
        \frac{\partial \mathcal{L}_{\text{phys}}}{\partial \mathbf{x}_t}
    \]
\end{enumerate}
If the molecule decoded at \(t = 0.99\) is not \textsc{RDKit}‑valid, no guidance is applied to the sample. 

\section{Experiments}

\subsection{Experimental Setup}

\paragraph{Training dataset}
We train \modelname{} on GEOM-Drugs \cite{axelrod2022geom}, a dataset of 1M high-quality conformers of drug-like molecules. We use the splits from \citet{vignac2023midi} and, following \citet{irwin2025semlaflow}, we discard molecules with more than 72 heavy atoms from the training dataset, accounting for 1\% of the data. During testing, we sample from the distribution of molecule sizes in the test set, which was left unchanged.

\paragraph{Evaluation Metrics}
We evaluate generated molecules on several metrics: (i)~\textbf{Validity}: Whether a molecule can be sanitized with \textsc{RDKit}, (ii)~\textbf{Novelty}: Whether the canonical SMILES of the molecule is not present in the training set, (iii)~\textbf{Diversity}: Tanimoto similarity of molecule fingerprints, (iv)~\textbf{Strain Energy} \cite{harris2023posecheck}: Energy of the molecule compared to low energy conformers, (v)~\textbf{Root Mean Square Deviation} (RMSD): When comparing molecules, averaged distance between the atoms of two molecules, (vi)~\textbf{\posebusters{}} \cite{buttenschoen2024posebusters}: Evaluates steric clashes, valid bond lengths and bond angles, double bond and aromatic ring flatness, and sufficiently low strain energy with respect to simulated conformers. We employ \posebusters{} as our main metric for measuring conformer quality, because its array of tests are designed to test for physical plausibility. A molecule is only considered \posebusters{}-valid if it passes all tests. In existing generative models for 3D molecule generation, most other metrics have been saturated \cite{irwin2025semlaflow}.

\paragraph{Training}

We train three \modelname{} models at three sizes: Both \modelname{}-mild (3.7M params.) and \modelname{}-hot (15M params.) were trained on two 80GB A100 GPUs for 36 hours at a learning rate of $0.001$. \modelname{}-spicy (59M params.) was trained on the same resources for 72 hours with a learning rate of $0.0005$ (see Appendix~\ref{app:model-hyperparams}). During training, we augment each batch with 8 random rotations of the same molecules to improve equivariance. We apply Exponential Moving Averaging (EMA) with decay strength $0.999$ to the model weights, which we ablate in Section~\ref{sec:ablations}. We compare our models against EQGAT-diff \cite{le2023eqgat}, FlowMol \cite{dunn2024flowmol}, SemlaFlow \cite{irwin2025semlaflow}, and ADiT \cite{joshi2025allatomdiffusiontransformer} (see Appendix~\ref{app:previous_work}).

\subsection{\modelname{} Achieves High Physical Quality}
\label{sec:main-results}

\begin{table*}[t]
  \caption{Results on GEOM‑Drugs. We generate 1,000 molecules for each method. $^\alpha$Due to computational constraints, we evaluate statistics on GEOM-Drugs on a random subset of 20K training molecules. 
}
  \centering
  \begin{adjustbox}{max width=\linewidth}
    \begin{tabular}{lccccccc}
      \toprule
      \textbf{Method}  & \textbf{\# Params.} & \textbf{Validity\,$\uparrow$} & \textbf{Novelty\,$\uparrow$} & \textbf{Diversity\,$\uparrow$} & \textbf{\posebusters{}\,$\uparrow$} & \textbf{Strain Energy\,$\downarrow$} & \textbf{Time\,$\downarrow$} (s) \\
      \midrule
      GEOM-Drugs$^\alpha$ & - & 1.0 & 0.0 & 0.90 & 0.94 & - & - \\
      \midrule
      % MiDi & & & & & &  \\
      EQGAT-diff & 12M & 0.94 & 0.94 & 0.90 & 0.84 & 360.19 & 4310.94 \\
      FlowMol & 4.3M & 0.81 & 0.81 & 0.91 & 0.64 & 34.20 & 362.22 \\
      SemlaFlow & 22M & 0.93 & 0.93 & 0.91 & 0.88 & 18.20 & 201.22 \\
      ADiT & 150M & 0.98 & 0.97 & 0.91 & 0.86 & 46.36 & 521.21 \\
    \midrule
        \modelname{}{-mild} & 3.7M & 0.95 & 0.93 & 0.89 & 0.85 & 21.32 & 5.9 \\
        \modelname{}{-hot} & 15M & 0.98 & 0.93 & 0.88 & 0.91 & 14.16 & 10.67\\
        \modelname{}{-spicy} & 59M & 0.97 & 0.90 & 0.89 & 0.92 & 15.07 & 19.77 \\
        \midrule
        \modelname{}{-spicy} w/ guidance & 59M & 0.97 & 0.92 & 0.89 & 0.94 & 19.23 & 131.80 \\
        \bottomrule
    \end{tabular}
  \end{adjustbox}
  \label{tab:geom-drugs}
\end{table*}

Our main results are shown in Table \ref{tab:geom-drugs}, example molecules are shown in Figure~\ref{fig:sampled-mols}. \modelname{}-spicy (59M), surpasses all prior methods in physical plausibility, raising the \posebusters{} validity from the previous state-of-the-art of 0.88 to 0.92 (see Figure~\ref{fig:pareto}). Interestingly, most of the gain is achieved by the 15M parameter \modelname{}-hot variant, with only modest improvements from further scaling to 59M, suggesting diminishing returns beyond this point. All variants maintain strong molecular diversity ($\sim$0.89), indicating that architectural simplifications do not compromise sampling breadth and generalisation. Earlier models such as FlowMol, which performed well on traditional metrics, show significantly lower physical validity (0.64), further highlighting the need for domain-aware evaluation such as \posebusters{}. Despite their simplicity, \modelname{} models generate molecules up to 100$\times$ faster than some prior baselines, offering a practical advantage for large-scale or iterative workflows. Finally, we find that guidance modestly improves \posebusters{} validity to 0.94, matching the training dataset, though at a 7$\times$ increase in compute cost, suggesting its use may be best reserved for high-value targets or post hoc filtering.

\subsection{The Effect of Sequential Ordering}
\label{sec:sequential-ordering}

\begin{figure*}
    \centering
    \includegraphics[width=0.45\linewidth]{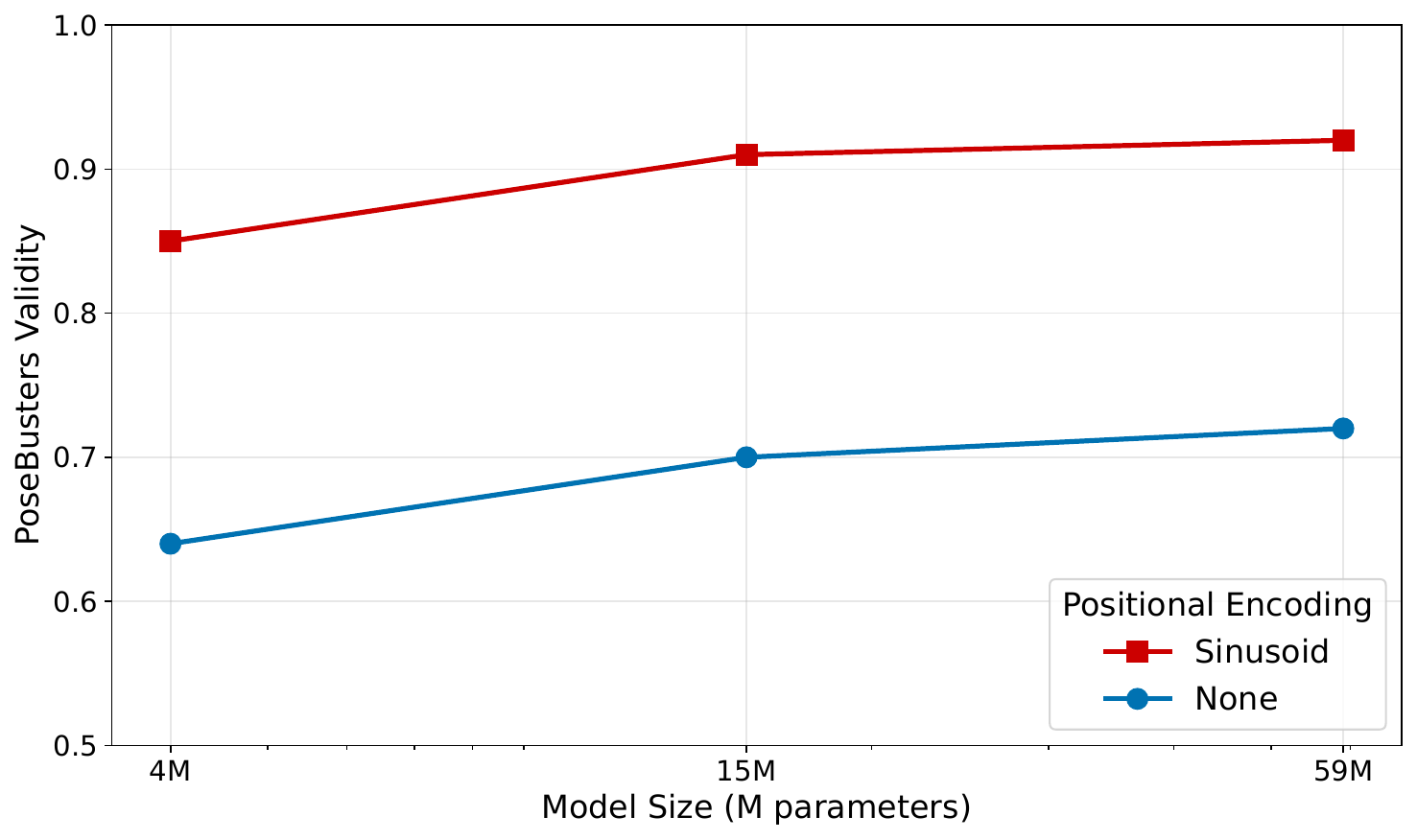}
    \includegraphics[width=0.45\linewidth]{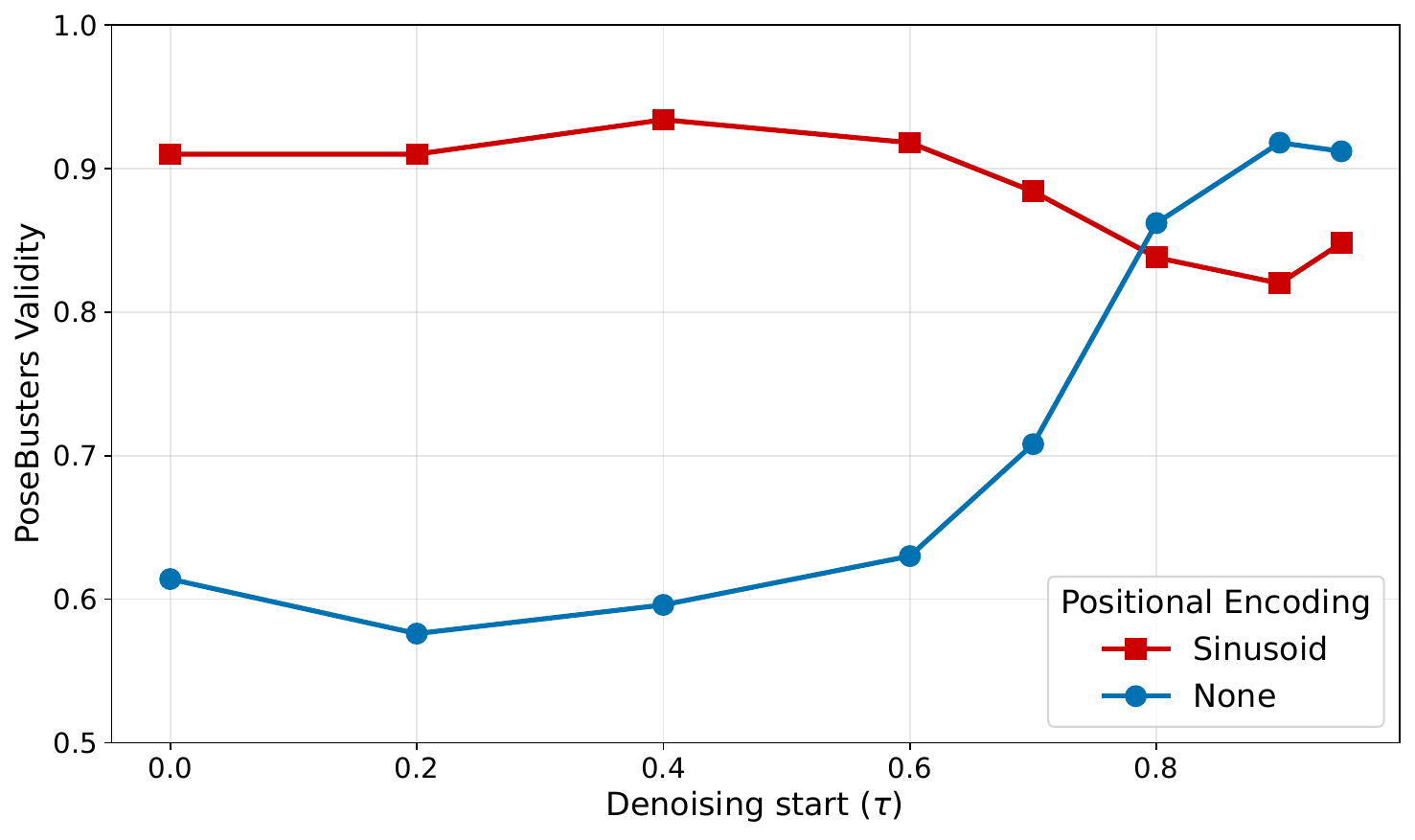}
    \caption{Left: Model performance across parameter scales with/without positional encodings. Right: 15M parameter model with/without positional encodings, \posebusters{} when starting denoising from different noise levels on test molecules.}
    \label{fig:sinusoid-v-bag}
\end{figure*}

\begin{figure*}[t]
    \centering
    \includegraphics[width=\linewidth]{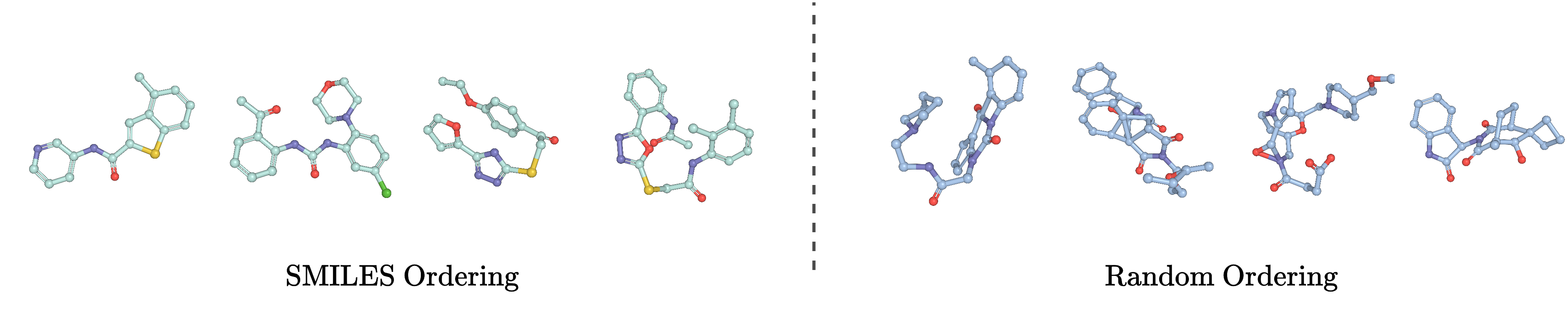}
    \caption{\textbf{Importance of atom ordering in \modelname{}}.
    Molecules generated with sinusoidal positional encodings from SMILES order (left) are coherent and valid, while random atom ordering (right) yields more fragmented, implausible structures, highlighting SMILES’ inductive bias for local structure during early denoising.
    }
    \label{fig:conformer-fail}
\end{figure*}

Empirically, we observe that introducing sequence positional encodings yields higher quality molecules compared to treating atoms in a bag-of-words fashion. We compare the effect of positional encodings across several model scales (see Figure~\ref{fig:sinusoid-v-bag}). We also show examples of failure modes we observed repeatedly in models without positional encodings in Figure~\ref{fig:conformer-fail}. We hypothesize that this difference in generative quality may stem from early steps in molecule denoising, when the atomic coordinates are very noisy and positional encodings can provide a signal about relative positions. To test this, we sample from two 15M parameter \modelname{}-models, one trained with sinusoid encodings and one without any encodings. We partially noise molecules to different $t\in[\tau, 1]$ and finish the denoising process with the models from that point. We do this to better isolate the sampling dynamics of the models at different noise levels. 

Figure~\ref{fig:sinusoid-v-bag} shows how as $\tau$ increases, the performance difference of the models decreases and switches near the end of denoising. This suggests that the sampling trajectories in the model with positional encodings differ from those the model is trained on (see Appendix~\ref{app:sampling}). Furthermore, the higher \posebusters{} validity of the positional-encoding-free model towards the end of denoising suggests that in later stages of denoising its sampling trajectory is well aligned with training trajectories. This indicates that it is able to create high quality molecules  when the final atom positions are apparent from its noisy coordinates, implying that as the final relative positions of atoms become more evident, positional encodings become less relevant.

\subsection{Ablations}
\label{sec:ablations}

Table \ref{tab:ablations} summaries how three “optional” components affect the 15M-parameter \modelname{}-hot model on GEOM-Drugs.
% EMA
Removing weight-EMA has almost no effect: all headline metrics change by $\leq0.01$ and diversity rises slightly. This shows that the model does not rely on EMA for chemical or geometric correctness.
% Cross atten
Performance is more sensitive to the coordination between input and output coordinate features. Eliminating the single cross-attention block lowers validity and novelty by $\sim$0.04 and, critically, drops \posebusters{} validity to 0.80. This indicates that coupling atom-type and coordinate information is necessary to resolve steric clashes and strain at this parameter scale.
% Pos en
Positional encodings also prove critical. Without them, raw validity remains high (0.93) but \posebusters{} validity collapses to 0.70, revealing widespread geometric artifacts. The model can still generate chemically plausible graphs, yet struggles to arrange them in physically realistic 3D space (see Figure~\ref{fig:conformer-fail}).
%Summary
In short, TABASCO’s competitive accuracy does not depend on heavy symmetry priors, but it does require sequence position cues and cross-attention with the latent inputs; the EMA has negligible impact on performance.
% Appendix
In Appendix~\ref{app:sampling} we further investigate the effect of removing random rotations alltogether, reducing the number of sampling steps and modifying the sampling strategy.

% ---- Final table ----
\begin{table*}[h]
  \centering
  \caption{Ablation study of model performance when removing components. Layer counts are adjusted to match model size where needed. Higher values are better.}
  % \begin{adjustbox}{max width=\linewidth}
    \begin{tabular}{lcccc}
      \toprule
      \textbf{Method} & \textbf{Validity} & \textbf{Novelty} & \textbf{Diversity} & \textbf{\posebusters{}} \\
      \midrule
        \modelname{}{-hot} & 0.98 & 0.93 & 0.88 & 0.91 \\
        \midrule
        w/o EMA & 0.98 & 0.93 & 0.89 & 0.91 \\
        w/o cross-attention & 0.94 & 0.89 & 0.89 & 0.80 \\
        w/o positional encoding & 0.93 & 0.93 & 0.91 & 0.70 \\
        \bottomrule
    \end{tabular}
    \label{tab:ablations}
  % \end{adjustbox}
\end{table*}

% \paragraph{The Effect of Cross-Attention}

% We study the effect of the final layer of cross-attention in our model (see Figure~\ref{fig:model-arch}). To this end, we train several comparison models, where compensate the final cross-attention layer with additional transformer blocks.

% \paragraph{Self-conditioning}

\subsection{Evaluating Equivariance}

\begin{figure}[t]
    \centering
    \includegraphics[width=0.5\textwidth]{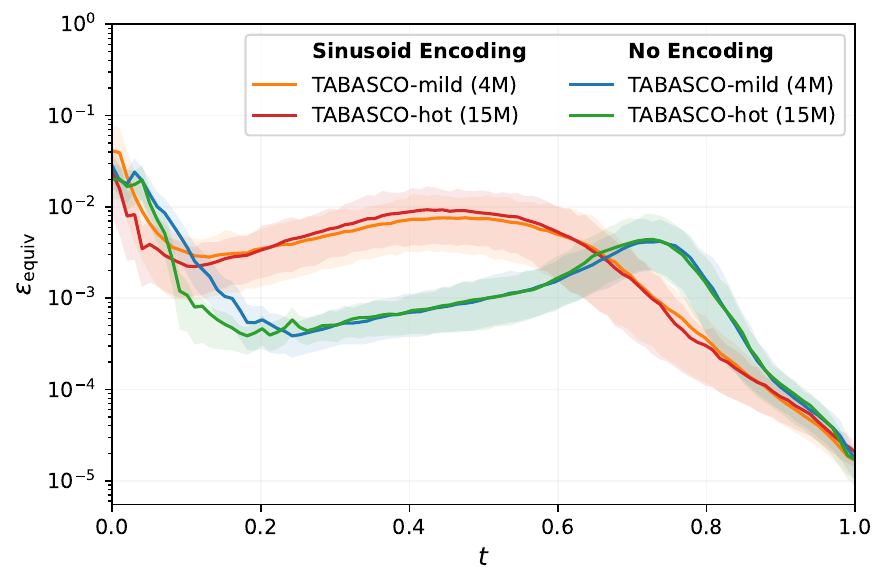}
    \caption{Relative equivariance error for TABASCO-hot (15M) and -mild (3.7M) with and without positional encodings. The error is normalized to the average atom coordinate magnitude.}
    \label{fig:equiv-error}
\end{figure}

We evaluate the quality of \modelname{}'s equivariance, as this is not encoded into the architecture. Similarly to previous work, we measure the deviation of the models prediction under random rotations \cite{karras2021equiverror, bouchacourt2021equiverror}. To control for numerical inaccuracies during sampling, rather than measuring the equivariance of fully denoised molecules, we measure the equivariance of the endpoint predictions of the model by partially noising molecules to different $t \in [0,1]$. Given noisy coordinates $\mathbf{x}_t$ at time $t$, a random rotation $R$, and a function that at any timestep predicts the endpoint $\hat{\mathbf x}_1 = f(\mathbf{x}_t, t)$, we randomly rotate the input and apply the inverse rotation to the output, i.e.  $Z(\mathbf{x}_t, t, R) = R^\top f(R\mathbf{x}_t, t)$. We estimate the relative equivariance error with

\begin{equation}
   \epsilon_\text{equiv} = \text{Var}_{R}\left[\frac{Z(\mathbf{x}_t, t, R)}{||Z(\mathbf{x}_t, t, R)||}\right]\;.
\end{equation}

We normalize the endpoint prediction per atom within random rotations of the same molecule to account for changes in scale during the sampling process and differing vector magnitudes. In an equivariant model one would have $Z(\mathbf{x}_t, t, R) = f(\mathbf{x}_t, t)$, which would trivially yield $\epsilon_\text{equiv} = 0$. In Figure~\ref{fig:equiv-error} we show that the relative equivariance error is small across all $t$, and decreases further as denoising progresses. We observe significant differences in the relative equivariance error when comparing models with and without positional encodings of up to an order of magnitude at different points in sampling.

\subsection{Effect of Physically-Constrained Guidance}

\begin{table*}[h]
  \caption{Effect of distance‑bounds guidance on \textsc{PoseBusters} validity and runtime (single A100) on \modelname{}-hot (15M) for 1000 molecules. RMSD is evaluated as a per-molecule mean with respect to the unguided baseline. $^\gamma$UFF calculations were performed on an M3 MacBook Pro.}
  \label{tab:phys_results}
  \centering
  \begin{adjustbox}{max width=\linewidth}
  \begin{tabular}{lccccc}
    \toprule
    \textbf{Method} & \textsc{PoseBusters}$\uparrow$ & Strain Energy $\downarrow$ & Diversity $\uparrow$ & RMSD & Runtime $\downarrow$ \\
    \midrule
    Baseline & 0.91 & 14.16 & 0.88 & - & 10.67\\
    w/ UFF & 0.94 & 4.74 & 0.88 & 0.226 & 14.21$^\gamma$ \\
    w/ Constr. UFF & 0.93 & 11.15 & 0.88 & 0.084 & 23.42$^\gamma$ \\
    \midrule
    w/ guidance & 0.94 & 19.23 & 0.89 & 0.132 & 75.65 \\
    \bottomrule
  \end{tabular}
  \end{adjustbox}
\end{table*}

In Table~\ref{tab:phys_results} we compare physically-constrained guidance to UFF relaxation of unguided molecules. Molecules are jointly denoised up to $t=0.99$. In one experiment we allow for unconstrained relaxation and in another introduce a movement constraint of $0.1$Å on each atoms original location. We choose $\alpha_{phys}=0.01$ in all experiments.\\
Table~\ref{tab:phys_results} shows how distance‑bounds guidance improves \textsc{PoseBusters} validity, while preserving diversity and slightly increasing strain energy. Although distance-bounds guidance increases sampling time due to sequential bound computation and backpropagation, overall sampling remains faster than in prior approaches. We also observe that the incremental nature of the method largely preserves atom positions compared to the unguided baselines. Finally, we argue that this refinement preserves diversity, because the molecular hypothesis is essentially fixed by \(t = 0.99\) and the model is only guided to create a lower-energy conformer of the same molecule. \\
The approach is model‑agnostic and applies to any diffusion‑ or flow‑based 3D generator that exposes gradients with respect to atom coordinates. Unlike force‑conditioned samplers such as DiffForce \cite{kulyte2024forcediff}, which back‑propagate full molecular‑mechanics gradients onto every atom at every reverse step, our method enforces pre‑tabulated element‑specific distance bounds and needs \emph{no} energy evaluation or additional learnable parameters. 

\subsection{Discussion}
\label{sec:disscussion}

Our findings align with recent trends toward simpler architectures, as seen in AlphaFold3's success without explicit symmetry constraints: Although SE(3) equivariance is often considered essential, our non-equivariant model learns equivariant representations up to small errors and achieves state-of-the-art performance on physical plausibility benchmarks, suggesting that enforced symmetries may be restrictive for some generation tasks. 
Our stripped-down architecture is easily extensible and only models coordinates and atom types explicitly. Conversely, omitting explicitly modelled bonds can limit conditioning when aiming to enforce valences or bond types \cite{peng2024pocketxmol}. 
Physically-constrained guidance is shown to be effective for improving physical plausibility with minimal modifications, however compared to traditional methods like UFF Relaxation, it remains more expensive and converges to higher strain energies. Nevertheless, unguided generation yields a ten-fold speed improvement compared to previous methods, potentially making practical applications like large-scale virtual screening more feasible in the future.

\section{Conclusion}
In this work we present \modelname{}, a non-equivariant generative model for 3D small molecule design that exhibits enhanced scalability and performance on physical plausibility compared to baselines. We study the importance of positional embeddings and atom ordering for small molecules, investigate the emergent equivariant properties of our model and the effects of scaling the model to large sizes. We note that minor effects emerged at scale unlike for previous work on other molecular modalities \citep{abramson2024alphafold3, geffner2025proteina}, and highlight the important design elements that are conducive to model performance. We hope that our model serves as a compelling example of how minimalist architectures can be effectively applied to molecular design and that our code base acts as an extensible tool for integration in drug-discovery workflows, through conditioned generation on relevant modalities or RL-based property optimization.

% \section{Societal Impact}

% Our model enables faster and more accessible generation of physically plausible molecular structures, supporting applications in drug discovery and computational chemistry. 

\section*{Acknowledgements}
We thank Simon Mathis, Chaitanya Joshi and Vladimir Radenkovic for their useful discussions and comments which improved the quality of the paper.
%discussions on equivariance, architectures and usefulness.

\bibliography{example_paper}
\bibliographystyle{icml2025}

%%%%%%%%%%%%%%%%%%%%%%%%%%%%%%%%%%%%%%%%%%%%%%%%%%%%%%%%%%%%%%%%%%%%%%%%%%%%%%%
%%%%%%%%%%%%%%%%%%%%%%%%%%%%%%%%%%%%%%%%%%%%%%%%%%%%%%%%%%%%%%%%%%%%%%%%%%%%%%%
% APPENDIX
%%%%%%%%%%%%%%%%%%%%%%%%%%%%%%%%%%%%%%%%%%%%%%%%%%%%%%%%%%%%%%%%%%%%%%%%%%%%%%%
%%%%%%%%%%%%%%%%%%%%%%%%%%%%%%%%%%%%%%%%%%%%%%%%%%%%%%%%%%%%%%%%%%%%%%%%%%%%%%%
\newpage
\appendix
\onecolumn
\section{Comparison to Previous Work}
\label{app:previous_work}

For all compared baselines we sample 1000 molecules with three random seeds on an A100 GPU. We report averages over the three runs.

\paragraph{EQGAT-diff}
We evaluated EQGAT-diff using the official codebase on GitHub\footnote{\nolinkurl{https://github.com/jule-c/eqgat_diff/}, available under the MIT License} and the checkpoints linked there. We used the example evaluation script, which we edited to save molecules as outputted from reverse sampling, without any post-processing. 

\paragraph{FlowMol}
We used the official implementation and the linked checkpoints on GitHub\footnote{\nolinkurl{https://github.com/Dunni3/FlowMol/}, available under the MIT License} and sampled molecules using the default script. For both GEOM\footnote{%
    Axelrod, Simon, et al.\ ``GEOM, energy-annotated molecular conformations
    for property prediction and molecular generation.'' \emph{Sci Data} 9, 185
    (2022). Available under CC0 1.0.
  } and QM9 \cite{ramakrishnan2014quantum} we benchmarked againts the CTMC-based models.

\paragraph{Semla Flow}
We evaluated SemlaFlow using the sampling script and model checkpoints from GitHub\footnote{\nolinkurl{https://github.com/rssrwn/semla-flow/}, available under the MIT License}. We modified the sampling script to save all outputs from the model, as opposed to only valid molecules. 

\paragraph{ADiT}
We benchmark ADiT by evaluating the molecules provided in the paper's GitHub repository\footnote{https://github.com/facebookresearch/all-atom-diffusion-transformer}.

\paragraph{Performance Comparison on QM9}
We further train and evaluate our model on the benchmark dataset QM9 \cite{ramakrishnan2014quantum} and compare the performance to previous methods in Table~\ref{tab:qm9-results}. We observe that \modelname{} achieves very low novelty scores on QM9 along with \posebusters{} close to one. Given that QM9 represents a subset of physically plausible molecules from an enumeration containing up to nine heavy atoms, the low novelty score is not surprising, and is a testament to the fact that our model's outputs are confined within the constraints of physical plausibility. 
We further note that on QM9 adding positional encodings still helps with performance, but the performance gap is much smaller compared to \modelname{} on GEOM-Drugs. A possible explanation for this is that the much smaller number of atoms per molecule compared to GEOM-Drugs makes it easier to distinguish atoms and place them with respect to each other even without positional encodings.

\begin{table}[h]
    \centering
    \caption{Results on the QM9 Dataset}
    \begin{adjustbox}{max width=\linewidth}
    \begin{tabular}{lcccccc}
      \toprule
      \textbf{Method}  & \textbf{\# Params.} & \textbf{Validity\,$\uparrow$} & \textbf{Novelty\,$\uparrow$} & \textbf{Diversity\,$\uparrow$} & \textbf{\posebusters{}\,$\uparrow$} & \textbf{Strain Energy\,$\downarrow$}\\
      \midrule
      EQGAT-diff & 12M & 0.99 & 0.99 & 0.89 & 0.94 & 9.10 \\
      FlowMol & 4.3M & 0.97 & 0.97 & 0.92 & 0.92& 17.81 \\
      SemlaFlow & 22M & 0.99 & 0.99 & 0.89 & 0.95 & 4.69 \\
      \midrule
      TABASCO-mild & 3.7M & 0.98 & 0.31 & 0.91 & 0.98 & 2.31 \\
      TABASCO-hot & 15M & 0.99 & 0.32 & 0.92 & 0.99 & 3.32 \\
      \hspace{1em}\textit{w/o pos. encodings} & 15M & 1.00 & 0.34 & 0.93 & 0.93 & 17.10 \\
      \bottomrule
    \end{tabular}
    \end{adjustbox}
    \label{tab:qm9-results}
\end{table}

% We downloaded the code and the GEOM-DRUGS checkpoint as given in \href{https://github.com/rssrwn/semla-flow}{\texttt{github.com/rssrwn/semla-flow}} and sampled 1000 molecules. Upon inspection of the code, we found that the SemlaFlow sampling script did not save invalid molecules\footnote{See code at: https://github.com/rssrwn/semla-flow/blob/main/semlaflow/predict.py\#L224}, thus artificially inflating PoseBusters validity in our downstream benchmarking. Hence, modified this code to also take into account invalid molecules. 

\section{Further Ablations}
\label{app:sampling}

\paragraph{Time Distribution During Training}

Based on success in previous works, we choose the Beta-distribution for sampling the time $t$ during training \cite{irwin2025semlaflow, geffner2025proteina}. We investigate the effect of different $\alpha$ values for the training time distribution $\text{Beta}(\alpha,1)$ and ablate three values in Table~\ref{tab:train-time-alpha}. We observe significant changes in performance at sampling time when shifting the probability weight assigned to different times $t\in[0,1]$ during training, and empirically find that $\text{Beta}(1.8, 1)$ yields the best results for our purpose.

\begin{table}[]
    \centering
    \caption{Ablation of $\alpha$ in the training time $t$ distribution $\text{Beta}(\alpha,1)$ on \modelname{}-hot (15M) trained on GEOM-Drugs.}
    \begin{tabular}{lcccc}
        \toprule
        \textbf{$\alpha$} & \textbf{Validity\,$\uparrow$} & \textbf{Novelty\,$\uparrow$} & \textbf{Diversity\,$\uparrow$} & \textbf{\posebusters{}\,$\uparrow$} \\
        \midrule
        1.5 & 0.96 & 0.92 & 0.89 & 0.84 \\
        1.8 & 0.98 & 0.93 & 0.88 & 0.91 \\
        2.0 & 0.97 & 0.93 & 0.88 & 0.89 \\
        \midrule
    \end{tabular}
    \label{tab:train-time-alpha}
\end{table}

\begin{figure}
    \centering
    \includegraphics[width=\linewidth]{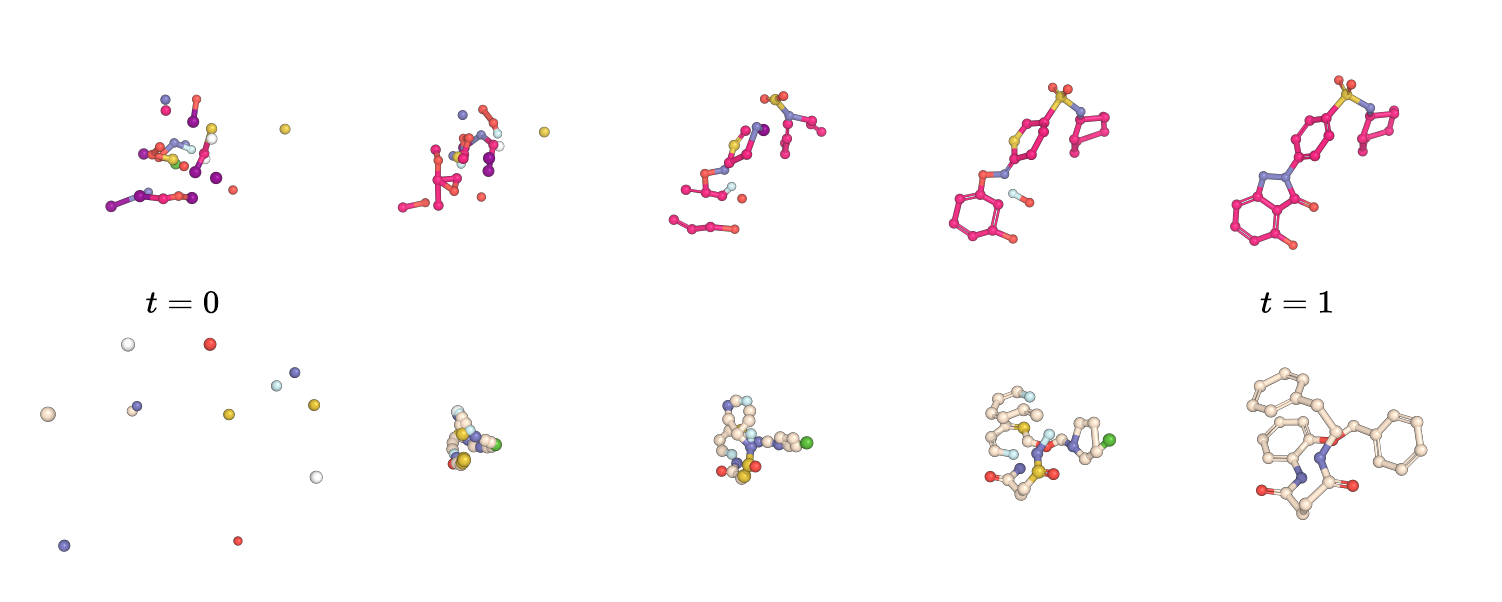}
    \caption{Snapshots of the sampling trajectories for two different molecules sampled from \modelname{}-hot (15M) trained on GEOM-Drugs. The upper trajectory is sampled with $g(t)=0$ and the lower one with $g(t)=\frac{1}{t+0.01}$}
    \label{fig:two-trajectories}
\end{figure}

\paragraph{Stochasticity Ablation}
Based on the approach in \cite{geffner2025proteina}, we investigate several choices for $g(t)$ in Eq. \ref{eq:stochastic_sampling}. Throughout this work, except when explicitly stated otherwise, we set $g(t)$ to zero for $t>0.9$ to allow for precise placement of atoms towards the end of sampling. Table~\ref{tab:stochasticity-ablation} compares the effect of four possible stochasticity functions. We observe that except for $g(t)=0$ performance is very similar across all metrics. We trace the contrast between $g(t)=0$ and the other parameterizations to a difference in sampling trajectories and show a comparison in Figure~\ref{fig:two-trajectories}. In contrast to the trajectory of $g(t)=0$ which is consistent with the training trajectory, the rest of the $g(t)$ functions have very large magnitudes close to $t=0$, which empirically leads first to an explosion and then to a collapse of the atom vector magnitudes. In the collapsed state atoms are roughly arranged in a sequence and slowly grow into the finished molecule as $t\rightarrow 1$ (see Figure~\ref{fig:two-trajectories}). This sudden rearrangement and growing into the finished molecule appears to yield better final molecules compared to when following the training trajectories more closely. This may also help explain the dip in \posebusters{} validity during partial molecule noising of \modelname{} with positional encodings in Figure~\ref{fig:sinusoid-v-bag}.
In contrast to this, we observe in Table~\ref{tab:stochasticity-ablation} that this explosion and collapse behaviour leads to much worse molecules when positional encodings are not added to the model, possibly because in the collapsed state atom coordinates are almost identical and they become very hard to distinguish without the positional encodings.

\begin{table}[]
    \centering
    \caption{Effect of four possible $g(t)$ parameterizations on \modelname{} trained on GEOM-Drugs. For all $t > 0.9$ we set $g(t)=0$ and use $\epsilon=0.01$.}
    \renewcommand{\arraystretch}{1.2}
    \begin{tabular}{ccccc}
        \toprule
        $g(t)$ & \textbf{Validity\,$\uparrow$} & \textbf{Novelty\,$\uparrow$} & \textbf{Diversity\,$\uparrow$} & \textbf{\posebusters{}\,$\uparrow$} \\
        \midrule
        $0$ & 0.96 & 0.95 & 0.90 & 0.83 \\
        $\frac{1}{t+\epsilon}$ & 0.98 & 0.93 & 0.88 & 0.91 \\
        $\frac{1}{t^2+\epsilon}$ & 0.97 & 0.93 & 0.89 & 0.91 \\
        $\frac{1-t}{t+\epsilon}$ & 0.98 & 0.94 & 0.89 & 0.89 \\
        \midrule
        \multicolumn{5}{c}{\textit{w/o positional encodings}} \\
        \midrule
        $0$ & 0.94 & 0.93 & 0.91 & 0.69 \\ 
        $\frac{1}{t+\epsilon}$ & 0.89 & 0.87 & 0.91 & 0.26 \\ 
        \bottomrule
    \end{tabular}
    \label{tab:stochasticity-ablation}
\end{table}

\paragraph{Number of Steps at Sampling}
We investigate the effect of reducing the number of sampling steps on molecule quality and ablate over several choices in Table~\ref{tab:num-sample-steps}. We observe that as little as 40 steps are necessary for \modelname{}-hot to outperform previous methods on \posebusters{}. We further observe that additional steps have no effect on molecular quality.

\begin{table}[]
    \centering
    \caption{Number of steps at sampling for \modelname{}-hot (15M) trained on GEOM-Drugs. We additionally evaluate connectivity, which denotes the fraction of fully connected molecules.}
    \begin{tabular}{lcccc}
        \toprule
        \textbf{\# Steps} & \textbf{Validity\,$\uparrow$} & \textbf{Novelty\,$\uparrow$} & \textbf{Connectivity\,$\uparrow$} & \textbf{\posebusters{}\,$\uparrow$} \\
        \midrule
        10 & 0.99 & 0.98 & 0.00 & 0.00 \\
        20 & 1.00 & 0.99 & 0.00 & 0.00 \\
        30 & 0.98 & 0.97 & 0.99 & 0.81 \\
        40 & 0.98 & 0.94 & 0.99 & 0.91 \\
        50 & 0.99 & 0.95 & 1.00 & 0.91 \\
        100 & 0.98 & 0.94 & 1.00 & 0.91 \\
        200 & 0.98 & 0.93 & 1.00 & 0.89 \\
        500 & 0.98 & 0.94 & 1.00 & 0.91 \\
        \midrule
    \end{tabular}
    \label{tab:num-sample-steps}
\end{table}

\begin{figure}
    \centering
    \includegraphics[width=0.7\linewidth]{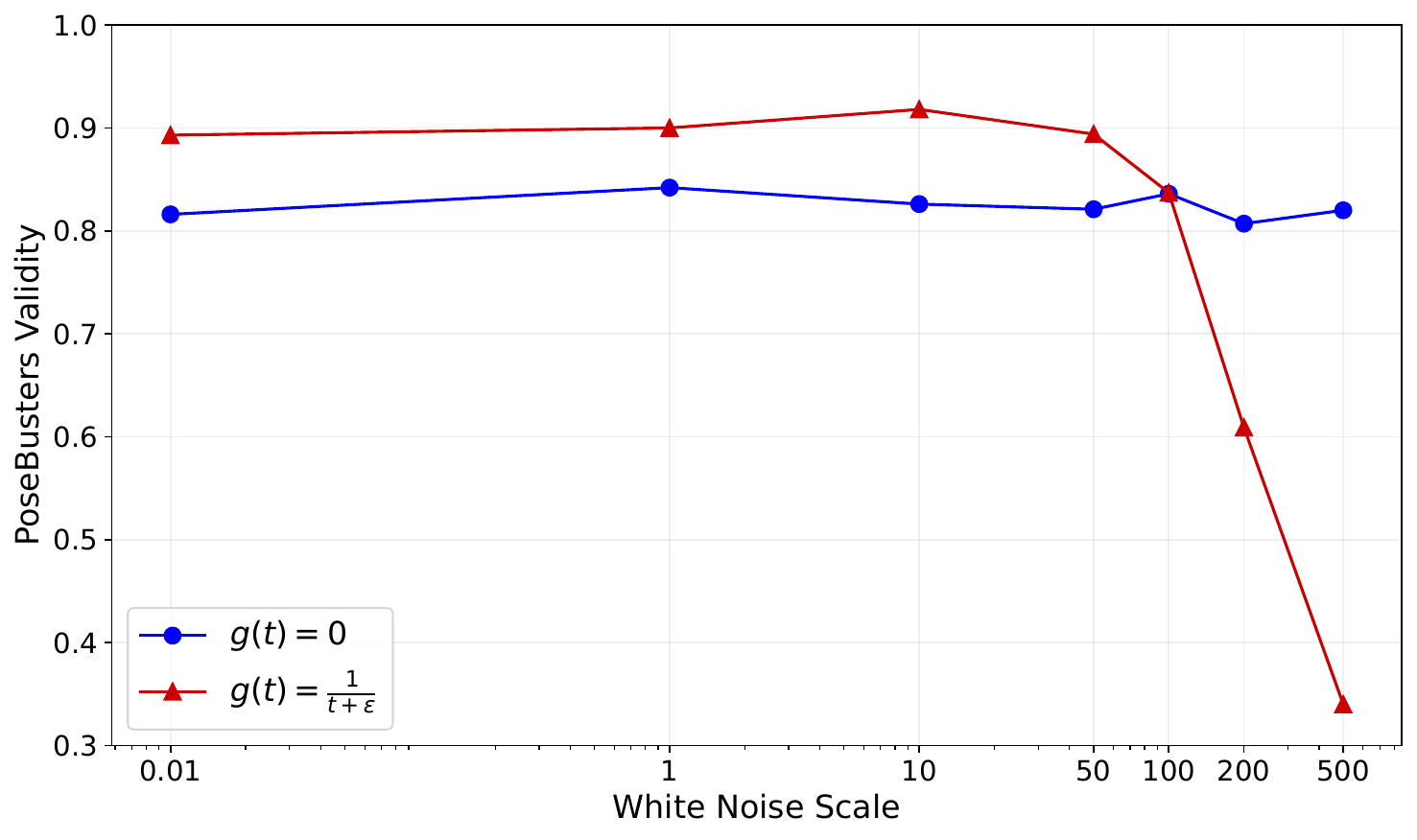}
    \caption{Comparison of \posebusters{}-validity across noise scales $\gamma$ with different $g(t)$. In contrast to all other comparisons we set $g(t)=0$ only beyond $t > 0.95$ to further augment the effect of adding noise.}
    \label{fig:noise-scale}
\end{figure}

\paragraph{Noise Scaling}
We investigate several values for $\gamma$ to ablate the effect of noise scaling. In Figure~\ref{fig:noise-scale} we compare \posebusters{}-validity for different noise scales and different $g(t)$ parameterizations. We observe that molecular quality remains high over several noise scales, and then collapses for $g(t)=\frac{1}{t+\epsilon}$.

\begin{figure}
    \centering
\includegraphics[width=0.8\linewidth]{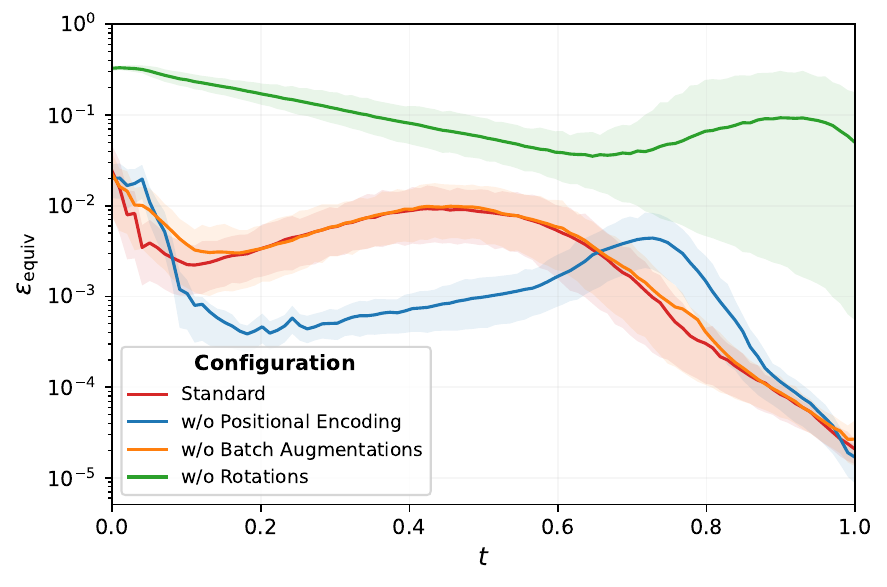}
    \caption{Comparison of the equivariance error to runs trained without per-batch random augmentations and without any random augmentation for \modelname{}-hot (15M) trained on GEOM-Drugs.}
    \label{fig:no-rots-equiv-err}
\end{figure}

\paragraph{The Effect of Random Rotations}
We study the effect of augmenting the data with random rotations during training on the sampled molecules. The batches in normal model training are augmented with seven copies of the same molecules (i.e. the effective batch size increases to eight times the original). All molecules in the augmented batch are then subjected individually to a random rotation. We study the effect of this operation by training two 15M \modelname{}-hot models, one where no augmentations and only random rotations are applied, and one where neither augmentations nor rotations are applied. To approximately match the original training dynamics, we increase the batch size of these models to match the effective batch size of the original training procedure. We train the models with the same compute budget as previously allotted: two A100 GPUs over 36 hours. We compare the observed performance in Table~\ref{tab:norot-noaug-perf} and visualize the equivariance error over time in Figure~\ref{fig:no-rots-equiv-err}. \\
The results show how omitting any random rotations of the data leads to a high-performing model that has a significantly higher equivariance error than all other models. Simultaneously, randomly rotating data, but omitting intra-batch augmentations with further random rotations, does not worsen the equivariance error, but slightly hurts \posebusters{} performance. This suggests that random intra-batch augmentations do not improve model equivariance, but improve training, possibly because of higher-quality gradient steps induced by the random rotations. Finally, we verify that random rotations are also not strictly necessary to create models that generate high-quality molecules, though at the expense of a significantly worse equivariance error. 

\begin{table}[]
    \centering
    \caption{Performance comparison of additional runs trained on GEOM-Drugs without per-batch random rotations and without any random rotation.}
    \begin{tabular}{lcccc}
    \toprule
        \textbf{Configuration} & \textbf{Validity\,$\uparrow$} & \textbf{Novelty\,$\uparrow$} & \textbf{Diversity\,$\uparrow$} & \textbf{\posebusters{}\,$\uparrow$} \\
        \midrule
        \modelname{}-hot (15M) & 0.98 & 0.93 & 0.89 & 0.91 \\
        \midrule
        w/o positional encoding & 0.93 & 0.93 & 0.91 & 0.70 \\
        w/o batch augmentations & 0.98 & 0.94 & 0.88 & 0.89 \\
        w/o random rotations & 0.98 & 0.94 & 0.89 & 0.90 \\
        \midrule
    \end{tabular}
    \label{tab:norot-noaug-perf}
\end{table}

\section{Extended Details on Models}
\label{app:model-hyperparams}

\begin{table}[htbp]
    \centering
    \caption{Model hyperparameters across different model sizes}
    \begin{tabular}{lccc}
    \toprule
    \textbf{Hyperparam.} & \textbf{\modelname{}-mild} & \textbf{\modelname{}-hot} & \textbf{\modelname{}-spicy} \\
    \midrule
    \# Params. & 3.711.369 & 14.795.529 & 59.082.249 \\
    Hidden size & 128 & 256 & 512 \\
    \# Transformer blocks & 16 & 16 & 16 \\
    \# Attn. heads & 8 & 8 & 8 \\
    Train $t$ distribution & Beta(1.8,1) & Beta(1.8,1) & Beta(1.8,1) \\
    $\lambda_\text{discrete}$ & 0.1 & 0.1 & 0.1 \\
    Learning rate & 0.001 & 0.001 & 0.0005 \\
    Optimizer & Adam & Adam & Adam \\
    EMA-weight & 0.999 & 0.999 & 0.999 \\
    \midrule
    Batch size & 256 & 256 & 128 \\
    \# Rotation Augs. & 8 & 8 & 8 \\
    Effective batch size & 2048 & 2048 & 1024 \\
    \midrule
    \# GPUs & 2 & 2 & 2 \\
    Training Duration & 36h & 36h & 72h \\
    \midrule
    \# Sampling Steps & 100 & 100 & 100 \\
    $g(t)$ & $\frac{1}{t + 0.01}$ & $\frac{1}{t + 0.01}$ & $\frac{1}{t + 0.01}$ \\
    $\gamma$ & 0.01 & 0.01 & 0.01 \\
    \bottomrule
    \end{tabular}
    \label{tab:model_hyperparams}
\end{table}

We give an overview of key design features of \modelname{} in Figure~\ref{fig:model-arch} and describe the unconditional sampling algorithm in detail in Algorithm~\ref{algo:standard-sampling}. In this section we further elaborate on model architecture and give concrete values for relevant hyperparameters in Table~\ref{tab:model_hyperparams}. \\
Atom coordinates are encoded with a bias-free linear layer that scales to the model's hidden size. Discrete atom types are encoded through an embedding layer, where we model Carbon, Nitrogen, Oxygen, Fluorine, Sulfur, Chlorine, Bromine, Iodine and a miscellaneous "$*$" atom, for all elements in the training set not contained within the previous list. We encode the time $t\in[0,1]$ through a Fourier encoding, and each atoms location within the molecules's SMILES sequence through a standard sinusoid encoding. We tried concatenating these four vectors and creating a combined hidden representation with an MLP mapping from $\mathbb{R}^{4\,\times\,\text{hidden dim}} \rightarrow \mathbb{R}^{\text{hidden dim}}$ but observed no difference in practice to simply adding the vector representations, and thus opted for this approach. Each transformer block applies layer-norm to the activations, then PyTorch multi-head attention, another layer-norm and a transition layer, where we include residual connections between the first two and second two components. This output is processed by two parallel PyTorch cross-attention layers one for atom types and for coordinates. Each consists of a self-attention block, a multi-head attention block, where the original combined hidden representation is used as key and value, and a feed-forward block. Both outputs are subsequently processed through domain-specific MLPs where the output atom coordinate MLP is also bias-free.

\section{Further Analysis on Physical Guidance}
\label{app:phys-guidance}

\begin{table}[h]
  \caption{Comparison \posebusters{} validity when adding physically-constrained guidance. Evaluated on 1000 molecules on a single A100 GPU.}
  \label{tab:all-phys-guidance}
  \centering
  \begin{adjustbox}{max width=\linewidth}
  \begin{tabular}{lccccccc}
    \toprule
    \textbf{Method}  & \textbf{\# Params.} & \textbf{Validity\,$\uparrow$} & \textbf{Novelty\,$\uparrow$} & \textbf{Diversity\,$\uparrow$} & \textbf{\posebusters{}\,$\uparrow$} & \textbf{Strain Energy\,$\downarrow$} & \textbf{Time\,$\downarrow$} (s) \\
    \midrule
    \modelname{}{-mild} & 3.7M & 0.95 & 0.93 & 0.89 & 0.85 & 21.32 & 5.9 \\
    \hspace{1em}\textit{w/ guidance} & & 0.96 & 0.95 & 0.89 & 0.91 & 26.53 & 60.86 \\
    \midrule
    \modelname{}{-hot} & 15M & 0.98 & 0.93 & 0.88 & 0.91 & 14.16 & 10.67 \\
    \hspace{1em}\textit{w/ guidance} & & 0.97 & 0.94 & 0.89 & 0.94 & 19.23 & 75.66 \\
    \midrule
    \modelname{}{-spicy} & 59M & 0.97 & 0.90 & 0.89 & 0.92 & 15.07 & 19.77 \\
    \hspace{1em}\textit{w/ guidance} & & 0.97 & 0.93 & 0.89 & 0.94 & 17.01 & 131.80 \\
    \bottomrule
  \end{tabular}
  \end{adjustbox}
\end{table}

We provide a detailed description of our physically constrained guidance procedure in Algorithm~\ref{algo:phys-guidance} and provide full results on all model sizes in Table~\ref{tab:all-phys-guidance}. We observe the largest improvement in \posebusters{} for \modelname{}-small, and minor improvements in novelty for all model sizes. Simultaneously we consistently observe an increase in strain energy when applying physical guidance.

\section{Limitations}
\label{app:limitations}
The approach described in this work introduces several limitations. SMILES-derived positional encodings improve performance but can theoretically introduce systemic biases, that may limit the model when faced with unusual bond patterns or non-standard chemical structures. Furthermore, omitting explicit bond modeling creates a leaner model and simpler training objective, but limits control over valences and bond orders when sampling the model. While \modelname{} exhibits emergent equivariance, in some areas such as molecular dynamics, where even small equivariance errors can prove problematic, this approximate equivariance may still be insufficient.\\
The physically-constrained guidance algorithm serves as a proof-of-concept for boosting the physical plausibility of molecules during sampling without requiring any modifications to training data, model architecture or parameter scale. Still, this approach is based on optimizing chemoinformatics heuristics for high-quality molecules and it dramatically increases sampling times.\\
Furthermore, while useful to quantify physical plausibility of 3D molecules, \posebusters{} cannot capture all aspects of molecular quality, and does not quantify additional very relevant metrics of interest: \modelname{} does not address improvements in drug-likeness of molecules or synthetic accessibility.

\begin{algorithm}
    \caption{Unconditional Sampling Algorithm}
    \label{algo:standard-sampling}
    \begin{algorithmic}[0]
        \Procedure{EuclideanStep}{$\mathbf{x}_t, \hat{\mathbf{x}}_1, t, \Delta t, g(\cdot), \gamma$}
        \State $\mathbf{v}_t \gets \frac{1}{1-t}(\hat{\mathbf{x}}_1 - \mathbf{x}_t)$
        \State $\mathbf{s}_t \gets g(t)\frac{t\mathbf{v}_t - \mathbf{x}_t}{1-t}$
        \State $\text{d}W_t \gets \sqrt{2\,\gamma\,g(t)}\; \mathcal{N}(0, I)$
        \State $\mathbf{x}_t \gets (\mathbf{v}_t + \mathbf{s}_t + \text{d}W_t) \Delta t$
        \State \Return $\mathbf{x}_t$
        \EndProcedure
        \Statex
        \Procedure{DiscreteFlowStep}{$a_t, \hat{p}_1, t, \Delta t$} \Comment{All indexed ops without loops are vectorized}
        \State $\mathbf{r}_t(i, \cdot) = \frac{\Delta t}{1-t}\hat{p}_1(i)$ \Comment{$\hat{p}_1$ consists of softmax-normalized model logits}
        \State $\mathbf{r}_t(i, a_{t}(i)) \gets - \sum_{j\neq a_{t}(i)} \mathbf{r}_t(i, j)$ \Comment{Make $\mathbf{r}_t$ zero mean}
        \State $\mathbf{p}_{t + \Delta t}(i,j) \gets 1_{a_t(i)=j} + \mathbf{r}_t(i,j)$
        \State $a_{t}(i) \gets \text{Categorical}(\mathbf{p}_{t + \Delta t}(i,\cdot))$ 
        \State \Return $a_t$
        \EndProcedure
        \Statex
        \Procedure{SampleMolecule}{$f, \{t_i\}_{i=0}^{N}$}
            \State $\mathbf{x} \gets \mathcal{N}(0, I)$
            \State $a \gets \text{Categorical}\left(\delta(\frac{1}{\text{\# atom types}})\right)$
            \For{$i = 1$ \textbf{to} $N$}
                \State $\left.\begin{array}{l}
                    \Delta t \gets t_i - t_{i-1} \\
                    (\hat{\mathbf{x}}_1, \hat{p}_1) \gets \text{EndpointPrediction}(f, (\mathbf{x}, a), t_i) \\
                    \mathbf{x} \gets \text{\textsc{EuclideanStep}}(\mathbf{x}_t, \hat{\mathbf{x}}_1, t_i, \Delta t) \\
                    a \gets \text{\textsc{DiscreteFlowStep}}(a_t, \hat{p}_1, t_i, \Delta t)
                \end{array}\right\} \text{\textsc{SamplingStep}}$
            \EndFor
            \State \Return $(\mathbf{x}, a)$
        \EndProcedure
    \end{algorithmic}
\end{algorithm}

\begin{algorithm}
\caption{Flow Matching with Physical Guidance}
\label{algo:phys-guidance}
\begin{algorithmic}[1]
\Procedure{PhysicalGuidance}{$f, (\mathbf{x}_t, a_t), t, \alpha$}
    \State $\hat{\mathbf{x}}_1, \hat{p}_1 \gets \text{EndpointPrediction}(f, (\mathbf{x}_t, a_t), t)$ 
    \State $\hat{a}_1(i)=\text{argmax}_j\;\hat{p}_1(i, j)$
    \State $\text{bounds} \gets \text{GetPhysicalConstraints}(\hat{\mathbf{x}}_1, \hat{a}_1)$ \Comment{Calls \textsc{RDKit} \texttt{GetBoundsMatrix()}}
    \For{each atom pair $(\mathbf{x}_{t,i},\mathbf{x}_{t,j})$ in $\mathbf{x}_t$} \Comment{This nested loop is vectorized in practice}
        \State $d_{ij} \gets ||\mathbf{x}_{t,i}-\mathbf{x}_{t,j}||_2^2$ 
        \If{$d_{ij} < \text{bounds}_{ij}^{\min}$} \Comment{Can also regress towards the interval centre}
            \State $\mathcal{L} \gets \mathcal{L} + (d_{ij} - \text{bounds}_{ij}^{\min})^2$
        \ElsIf{$d_{ij} > \text{bounds}_{ij}^{\max}$}
            \State $\mathcal{L} \gets \mathcal{L} + (d_{ij} - \text{bounds}_{ij}^{\max})^2$
        \EndIf
    \EndFor
    \State $\mathbf{x}_t \gets \mathbf{x}_t - \alpha \cdot \text{sign}(\nabla_{\mathbf{x}_t} \mathcal{L})$ \Comment{The sign-op slightly stabilizes updates in practice}
    \State \Return $\mathbf{x}_t$
\EndProcedure
\Statex
\Procedure{GuidedSampling}{$f, (\mathbf{x}_0, a_0), \{t_i\}_{i=0}^{N}, t_{\text{guidance}}$}
    \State $(\mathbf{x}, a) \gets (\mathbf{x}_0, a_0)$
    \For{$i = 1$ \textbf{to} $N$}
        \State $\Delta t \gets t_i - t_{i-1}$
        \If{$t_i \geq t_{\text{guidance}}$}
            \State $\mathbf{x} \gets \text{\textsc{PhysicalGuidance}}(f, (\mathbf{x}, a), t_i, \alpha)$
        \EndIf
        \State $(\mathbf{x}, a) \gets \text{\textsc{SamplingStep}}(f, (\mathbf{x}, a), t_i, \Delta t)$
    \EndFor
    \State \Return $(\mathbf{x}, a)$
\EndProcedure
\end{algorithmic}
\end{algorithm}

%%%%%%%%%%%%%%%%%%%%%%%%%%%%%%%%%%%%%%%%%%%%%%%%%%%%%%%%%%%%%%%%%%%%%%%%%%%%%%%
%%%%%%%%%%%%%%%%%%%%%%%%%%%%%%%%%%%%%%%%%%%%%%%%%%%%%%%%%%%%%%%%%%%%%%%%%%%%%%%

\end{document}